\documentclass[letterpaper, 10 pt, conference]{ieeeconf}  % Comment this line out if you need a4paper

\IEEEoverridecommandlockouts                              % This command is only needed if 
\usepackage{amsmath}
\usepackage{amssymb}
\usepackage{amsfonts}
\usepackage{float}
\usepackage{graphics}
\usepackage{graphicx}
\usepackage{epsfig}
\usepackage{subcaption}
\usepackage{xcolor}
\usepackage{hyperref}
\DeclareMathOperator*{\argmax}{argmax}

\newcommand{\real}{\mathbb{R}}
\newcommand{\Mat}[1]{\mathbf{#1}}
\renewcommand{\Vec}[1]{\mathbf{#1}}

\newcommand{\transpose}{T}

\newcommand{\Exp}[1]{\text{Exp}({#1})}

\newcommand{\cost}{J}
\newcommand{\weight}{\mathbf{W}}

\newcommand{\Frame}[1]{\boldsymbol{\mathcal{F}}_{#1}}
\newcommand{\tf}{\mathbf{T}}
\newcommand{\trans}{\mathbf{r}}
\newcommand{\rot}{\mathbf{C}}
\newcommand{\Tf}[2]{\tf_{#1#2}}
\newcommand{\Trans}[3]{{}_{#1}^{}\trans_{#1#2}^{#3}}

\newcommand{\mytrans}[3]{\leftidx{_{#1}}\trans{\IfValueTF{#2}{_{#2#3\hspace{2pt}}}{}}}

\newcommand{\BodyTrans}[2]{{}_{#2}^{}\trans_{#1#2}}
\newcommand{\EstBodyTrans}[2]{{}_{#2}^{}\hat{\trans}_{#1#2}}
\newcommand{\Rot}[2]{\rot_{#1#2}}

\newcommand{\err}{\mathbf{e}}
\newcommand{\meas}{\tilde{\mathbf{z}}}
\newcommand{\mmodel}{\mathbf{h}}

\newcommand{\covar}{\mathbf{\Sigma}}
\newcommand{\error}{\Vec{e}}
\newcommand{\state}{\Vec{x}}
\newcommand{\State}[1]{\state_{#1}}
\newcommand{\otherState}{\boldsymbol\gamma}

\newcommand{\quat}{\mathbf{q}}
\newcommand{\Quat}[2]{\quat_{#1#2}}

\newcommand{\vel}{\mathbf{v}}
\newcommand{\Vel}[2]{{}_{#1}^{}\vel_{#1#2}}

\newcommand{\accel}{\mathbf{a}}

\newcommand{\bias}{\mathbf{b}}
\newcommand{\Bias}[1]{\bias_{#1}}

\newcommand{\EstQuat}[2]{\hat\quat_{#1#2}}
\newcommand{\calibParams}{\boldsymbol{\theta}}

\newcommand{\dalpha}{\delta{\boldsymbol{\alpha}}}

\title{\LARGE \bf Accurate and Interactive Visual-Inertial Sensor Calibration with Next-Best-View and Next-Best-Trajectory Suggestion}

\author{
    Christopher L. Choi$^{1}$, Binbin Xu$^{1}$ and Stefan Leutenegger$^{1, 2}$% <-this % stops a space
    \thanks{$^{1}$Smart Robotics Lab, Department of Computing, Imperial College London, United Kingdom.
            {\tt\small \{christopher.choi, b.xu17, s.leutenegger\}@imperial.ac.uk}}%
    \thanks{$^{2}$Smart Robotics Lab, Technical University of Munich, Germany.
            {\tt\small stefan.leutenegger@tum.de}}%
}

\begin{document}

\maketitle
\thispagestyle{empty}
\pagestyle{empty}

% ABSTRACT
\begin{abstract}
Visual-Inertial (VI) sensors are popular in robotics, self-driving vehicles, and augmented and virtual reality applications. In order to use them for any computer vision or state-estimation task, a good calibration is essential. However, collecting \textit{informative} calibration data in order to render the calibration parameters observable is not trivial for a non-expert. In this work, we introduce a novel VI calibration pipeline that guides a non-expert with the use of a graphical user interface and information theory in collecting \textit{informative} calibration data with Next-Best-View and Next-Best-Trajectory suggestions to calibrate the intrinsics, extrinsics, and temporal misalignment of a VI sensor. We show through experiments that our method is faster, more accurate, and more consistent than state-of-the-art alternatives. Specifically, we show how calibrations with our proposed method achieve higher accuracy estimation results when used by state-of-the-art VI Odometry as well as VI-SLAM approaches. The source code of our software can be found on: \url{https://github.com/chutsu/yac}.
\end{abstract}

% INTRODUCTION
\section{INTRODUCTION}

In order to use Visual-Inertial (VI) sensors in computer vision or state-estimation tasks the calibration parameters must first be obtained. Conventionally, VI sensors are calibrated by an expert who would often collect calibration data by positioning and moving the sensors in front of a calibration target such as a checkerboard or grid of fiducial markers, then use an offline calibration tool such as Kalibr~\cite{Furgale:etal:IROS2013} to estimate the sensor calibration parameters. Good calibration results, however, may only be achieved, if the right kind and right amount of data is collected. More specifically, two potential practical issues arise during data capture: first, the choice of calibration views and the range of motions needed is not immediately clear to the non-expert. Secondly, the amount of data the user has to collect for calibration is also unclear, often collecting too much or too little data. A common practice to address these issues is to collect \textit{multiple} calibration data sequences, however, this is impractical in the field and identifying which calibration is optimal becomes a tedious and time-consuming task.

A straight forward solution to this problem would be to mount the VI sensor on a robot arm and perform a rehearsed or optimal calibration ``dance'', such as in \cite{Ao:etal:ICRA2022}. However, this requires extra hardware and is not a practical solution for many applications.
As an alternative to classic offline calibration methods, one can estimate the calibration parameters within a state-estimation framework such as OKVIS~\cite{Leutenegger:etal:IJRR2014}, VINS-MONO~\cite{Qin:etal:TRO2018}, and OpenVINS~\cite{Geneva:etal:ICRA2020} in real-time. Note, however, that any of these frameworks require some form of sufficiently accurate initial calibration, as well as sufficient visual features and motion excitation, therefore suffering from similar issues as offline calibration. Furthermore, natural keypoints and the lack of precise knowledge of the corresponding 3D positions may not produce the best possible results.

\begin{figure}[tb]
  \centering
  \includegraphics[width=1.0\linewidth, trim={0 9cm 0 0cm},clip]{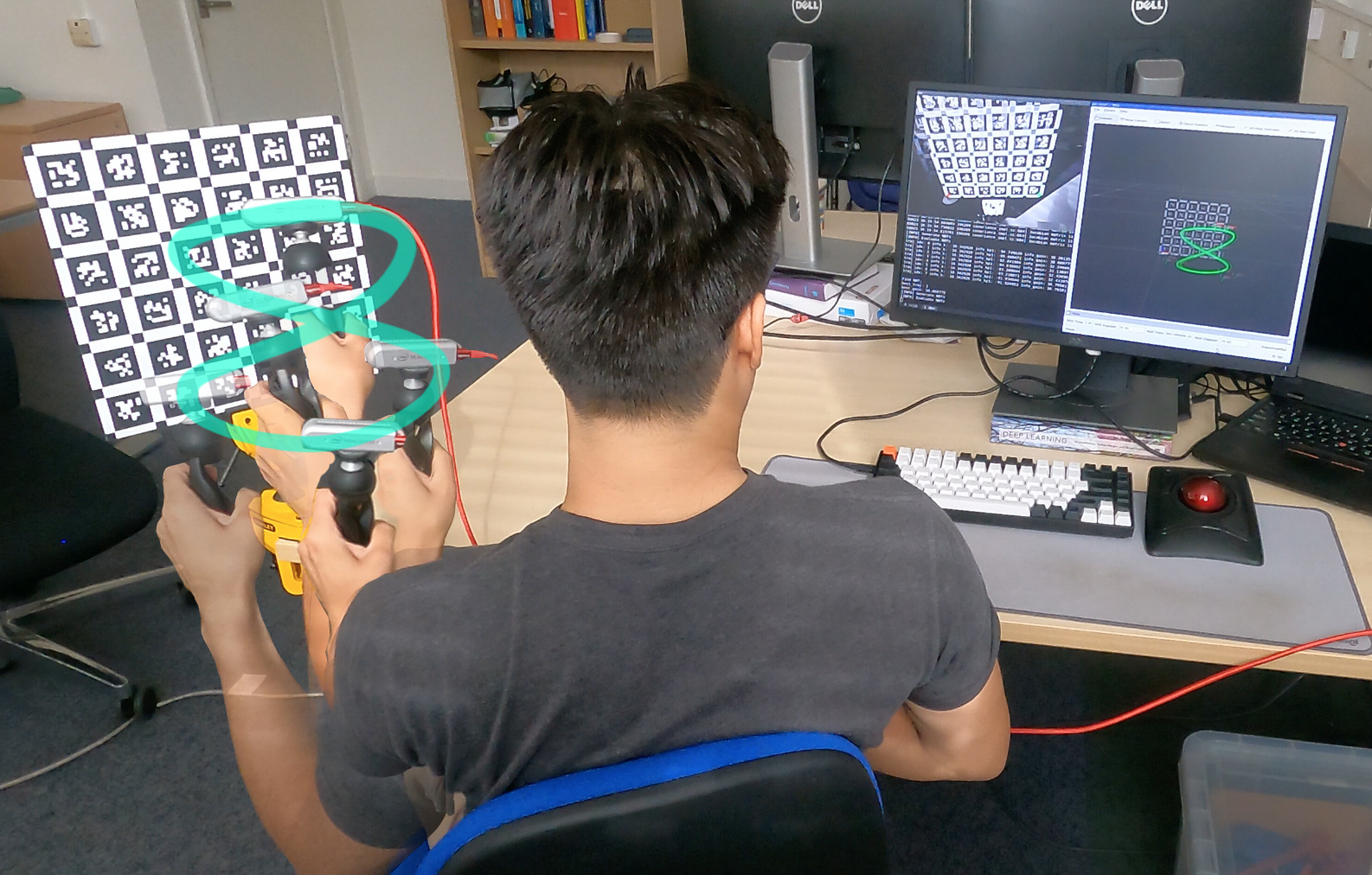}
  \caption{Our system interactively suggests next-best-actions to collect calibration data.}
  \label{fig:teaser}
  \vspace{-1.5em}
\end{figure}

In this work, we present an interactive VI sensor calibration pipeline that helps guide a non-expert in collecting \textit{informative} calibration data for a VI sensor \textit{once} through Next-Best-View (NBV) and Next-Best-Trajectory (NBT) suggestions (as shown in Fig.~\ref{fig:teaser}) in order to efficiently obtain sound calibrations. We show through extensive quantitative experiments on calibration sequences and several self-collected VICON real-world datasets that calibration parameters optimised through our system are more accurate and consistent than Kalibr by testing on state-of-the-art VI-SLAM ORBSLAM3~\cite{Campos:etal:TRO2021}.  In summary our contributions are:
\begin{itemize}
    \item{A complete and open-sourced interactive VI-camera calibration tool that supports any number of cameras; }
    \item{An information-theoretic procedure to identify the most informative Next-Best-View (NBV) and Next-Best-Trajectory (NBT) among a pre-defined set of viewpoints and trajectory primitives;}
    \item{An interactive graphical user interface for guiding the user through the calibration data collection process;}
    \item{Through experiments we show that our proposed method is faster, more accurate and more reliable compared to state of the art traditional \textit{non-guided} calibration methods, such as Kalibr~\cite{Furgale:etal:IROS2013}, even when used by \textit{novices}.}
\end{itemize}

% The remainder of this paper is organised as follows: we overview related work in Section~\ref{sec:related} and introduce the notation used in Section~\ref{sec:notation}. Then, we overview our system in Section~\ref{sec:overview}, which is later detailed concerning vision-only calibration with NBV in Section~\ref{sec:nbv} and VI extrinsics calibration and temporal calibration with NBT in Section~\ref{sec:nbt}. Section~\ref{sec:experiments} then presents the experimental results.

% RELATED WORK
\section{RELATED WORK}
\label{sec:related}

\textbf{Offline Methods}. In the robotics community, early works in VI-sensor calibration methods such as~\cite{Alves:etal:MIRC2003,Lobo:Dias:IJRR2007,Mirzaei:Roumeliotis:TRO2008,Dong:Mourikis:IROS2012} showed that it is possible to calibrate the extrinsics between a camera and IMU, with Kalibr~\cite{Furgale:etal:IROS2013} regarded as the current state-of-the-art tool. It is an offline method capable of calibrating a multi-camera system, as well as a VI system. However, the use of this tool requires expert knowledge, as the result is highly dependent on the quality of the calibration data captured. Therefore, the calibration process may in practice have to be repeated until desired results are reached.

\textbf{Online Methods}. State of the art state-estimation framework such as OKVIS~\cite{Leutenegger:etal:IJRR2014}, VINS-MONO~\cite{Qin:etal:TRO2018}, and OpenVINS~\cite{Geneva:etal:ICRA2020} can in practice estimate the calibration parameters in real-time. However, these frameworks require sufficiently accurate initial calibrations, as well as sufficient visual features and motion excitation in order to operate accurately.

\textbf{Reinforcement Learning Methods}: There has been a growing interest in using reinforcement learning for calibration such as~\cite{Ao:etal:ICRA2022,Chen:etal:CORL2021-calibration,Nobre:Heckman:IJRR2019} whereby the goal is to learn informative trajectories to render the VI-calibration parameters observable. However, the requirement of a robot arm to perform these motions is not always practical in the field. Further, these works do not provide quantitative results through a SLAM system to verify the optimality of the calibrated parameters.

\textbf{Information-Theoretic Methods}. The first calibration tool with an emphasis in guiding the user through capturing a good calibration sequence for a monocular camera is AprilCal~\cite{Richardson:etal:IROS2013}. The method used a quality metric to find and suggest the NBV in real-time during the camera calibration process. AprilCal, however, only supports calibrating the intrinsics of a single monocular camera. 

A more recent work that uses an information-theoretic approach for VI sensor calibration is the work of~\cite{Schneider:etal:ICRA2017,Schneider:etal:Sensor2019}, where they proposed a segment-based method for calibrating a VI sensor system in a AR / VR headset and self-driving car setting. The idea is to extract informative data during online state-estimation using an information-theoretic metric, and then perform a full-batch optimisation to update the calibration parameters offline. This approach, however, relies on the fact that the VI sensors are calibrated well initially. Secondly, the available data does not guarantee informative segments for calibration. 

In this paper, we place heavy emphasis on collecting \emph{informative} calibration data by using an information-theoretic metric to find the NBV and NBT in real-time, and by \emph{interactively} guiding the user in collecting them in order to calibrate the intrinsics, extrinsics, and time shift of a VI sensor. This is in contrast to current state-of-the-art calibration tools such as Kalibr~\cite{Furgale:etal:IROS2013} that assume the collected calibration data has sufficient views and range of motion.

% NOTATION
\section{NOTATION}
\label{sec:notation}

We employ the following notation throughout this work. Let $\Frame{W}$ denote the world reference frame. A 3D point $P$ in the world frame $\Frame{W}$ with respect to the origin is written as a position vector $\Trans{W}{P}{}$. A rigid body transformation from the body frame, $\Frame{B}$, to the world frame, $\Frame{W}$, is represented by a homogeneous transformation matrix, $\Tf{W}{B}$. Its rotation matrix component is written as, $\Rot{W}{B}$, and the corresponding Hamiltonian quaternion is written as, $\Quat{W}{B} = [ \boldsymbol{\eta}^{\transpose}, \epsilon ]^{\transpose} \in \mathcal{S}^{3}$, where $\epsilon$ and $\boldsymbol{\eta}$ are the real and imaginary parts. 

In general, the state vector we will be estimating lives on a manifold and thus we define an operator $\boxplus$ that will be used to perturb the states in tangent space such that $\state = \bar\state \boxplus \delta\state$, where $\bar\state$ is the state estimate and $\delta\state$ is the local perturbation. Vector quantities such as positions, velocities, biases are updated via standard vector addition. Rotation components on the other hand such as a quaternion are updated via a combination of the group operator $\otimes$ (quaternion multiplication) and exponential map $\Exp{\cdot}$, such that $\quat \boxplus \dalpha = \Exp{\dalpha} \otimes \quat$. As a result we will be using a minimal coordinate representation approach similar to~\cite{Leutenegger:etal:IJRR2014}. A comprehensive introduction to differential calculus is beyond of the scope of this paper, the reader is therefore encouraged to review~\cite{Bloesch:etal:CoRR2016,Sola:etal:ARXIV2018} for a more detailed treatment on the subject.

% BACKGROUND
\section{BACKGROUND}
\label{sec:background}

In robotics, the maximum a posteriori (MAP) estimator is commonly used to solve the camera and VI calibration problem,
\begin{equation}
    \label{eq:mle}
    \hat{\state} = \argmax_{\state} \; p(\state | \Vec{z}),
\end{equation}
where $\state$ is the state vector which may be comprised of poses, velocities, IMU biases and calibration parameters we are interested in jointly estimating, given the measurements, $\Vec{z}$. Assuming Gaussian measurements resulting in independent error terms $\Vec{e}_i$, maximising Eq.~\eqref{eq:mle} is equivalent to solving the sum of nonlinear least squares using a nonlinear optimisation algorithm such as the Gauss Newton method,
\begin{equation}
    \label{eq:gauss_newton}
    \sum_i\Mat{E}_i^{\transpose} \Mat{W}_i \Mat{E}_i \; \Delta {\state} = \sum_i
    -\Mat{E}_i^{\transpose} \Mat{W}_i \Vec{e}_i({\state}),
\end{equation}
where $\Delta{\state}$ is the update vector, $\Vec{e}_i({\state})$ is the $i^\mathrm{th}$ error term evaluated at the current estimate $\state$, $\Mat{E}_i$ is the Jacobian matrix of the error term and $\Mat{W}_i$ the measurement information. 

At convergence of the optimisation, we may approximate the posterior distribution as a Gaussian with mean $\state$ and find the covariance matrix $\hat{\covar}_{\state}$ by inverting the quantity $\sum_i\Mat{E}_i^{\transpose} \Mat{W}_i \Mat{E}_i$, also known as the Fisher Information matrix. However, recall that the state vector $\state$ not only contains calibration parameters $\calibParams$, but also other state variables not related to the calibration parameters which we denote as $\otherState$. In the context of calibration, we are only interested in the estimated calibration parameters, ${\calibParams}$, and the covariance of the calibration parameters ${\covar}_{\calibParams\calibParams}$. Expressing ${\state}$ and ${\covar}_{\state}$ in partition form,
\begin{equation}
    \label{eq:partition_blocks}
    {\state} = \begin{bmatrix}
        {\calibParams} \\
        {\otherState}
    \end{bmatrix}
    ,\enspace
    {\covar}_{\state} = \begin{bmatrix}
        {\covar}_{\calibParams\calibParams} & {\covar}_{\calibParams\otherState} \\
        {\covar}_{\otherState\calibParams} & {\covar}_{\otherState\otherState}
    \end{bmatrix},
\end{equation}
we can employ marginalisation on Normal distributions to get $p(\calibParams | \Vec{z}) = \mathcal{N}({\calibParams}, \covar_{\calibParams\calibParams})$, by extracting the corresponding blocks in Eq.~\eqref{eq:partition_blocks}. %In practice we will equivalently apply the Schur Complement on the Information Matrix.

\begin{figure*}[tb]
  \vspace{1em}
  \centering
  \includegraphics[width=\linewidth]{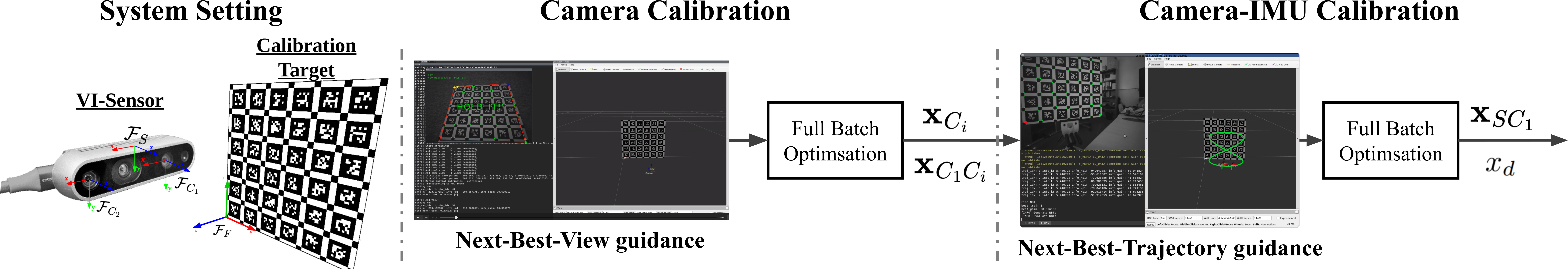}
  \caption{An overview of our VI calibration pipeline.}
  \label{fig:overview}
\end{figure*}

To objectively quantify whether the next VI measurements are informative for the VI calibration problem, we used the Mutual Information (MI) defined in~\cite{Maye:etal:IV2013},
\begin{align}
    \label{eq:mi}
    I(\calibParams_1 ; \meas_2) &=  
        \dfrac{1}{2} 
        \log 
        \dfrac{|
            \covar_{\calibParams_1 \calibParams_1}
        |}
        {| 
            \covar_{\calibParams_1 \calibParams_1| \Vec{z}_2}
        |},
\end{align}
where $\covar_{\calibParams_1 \calibParams_1}$ is the covariance estimate of $\calibParams$ using measurements $\Vec{z}_1$ alone, and $\covar_{\calibParams_1 \calibParams_1 | \Vec{z}_{2}}$ is the covariance estimate of $\calibParams$ using measurements $\Vec{z}_1$ and $\Vec{z}_2$, finally $|\cdot|$ is the matrix determinant. In summary, with Eq.~\eqref{eq:mi} we can measure the amount of information $\Vec{z}_2$ (next VI-sensor measurements) conveys to our current estimate $\calibParams | \Vec{z}_{1}$.

% SYSTEM OVERVIEW
\section{SYSTEM OVERVIEW}
\label{sec:overview}
An overview of our proposed calibration system is illustrated in Fig.~\ref{fig:overview}\footnote{The pipeline is demonstrated in details in the supplementary video.}. It consists of two stages. The first stage aims to perform vision-only camera intrinsics and extrinsics calibration employing Next-Best-View (NBV) feedback. In the second stage the camera-IMU extrinsics are found by using Next-Best-Trajectory (NBT) feedback, with the camera intrinsics and extrinsics obtained in the previous stage fixed. Both stages of the calibration process require the use of a static fiducial marker grid of known size as a calibration target. Specifically, we use a planar calibration target grid of AprilTags~\cite{Richardson:etal:IROS2013} introduced by Kalibr~\cite{Furgale:etal:IROS2013}. Throughout this work, the VI sensor to be calibrated is assumed to capture images and inertial measurements with the same clock source.

% CAMERA INTRINSICS AND EXTRINSICS CALIBRATION
\section{Camera Intrinsics and Extrinsics Calibration}
\label{sec:nbv}

In the following, we detail our approach of using Mutual Information (MI) and Next-Best-View (NBV) to calibrate intrinsics and extrinsics of all cameras.

\subsection{States}
\label{subsec:calib_camera-states}

For the camera calibration problem, the states to be estimated consist of the camera poses relative to the fiducial target coordinate frame $\Frame{F}$ as $\state_{FC_{1}}$, camera extrinsics relative to reference camera 1, $\state_{C_{1}C_{i}}$, and camera intrinsics, $\State{C_i}$, of the form:
\begin{align}
    % Camera pose
    \State{FC_{1}} &= \begin{bmatrix}
        \Trans{F}{C_{1}}{\transpose} 
        \;\; \Quat{F}{C_{1}}^{\transpose}
    \end{bmatrix}^{\transpose} 
    \in \real^{3} \times \mathcal{S}^{3}, \nonumber \\
    % Camera-camera extrinsics
    \State{C_{1}C_{i}} &= \begin{bmatrix}
        \Trans{C_{1}}{C_{i}}{\transpose} 
        \;\; \Quat{C_{1}}{C_{i}}^{\transpose}
    \end{bmatrix}^{\transpose} 
    \in \real^{3} \times \mathcal{S}^{3}, \\
    % Camera intrinsics
    \State{C_i} &= \begin{bmatrix}
       f_{x} 
       \;\; f_{y} 
       \;\; c_{x} 
       \;\; c_{y} 
       \;\; k_{1} 
       \;\; k_{2} 
       \;\; p_{1} 
       \;\; p_{2}
    \end{bmatrix}^{\transpose} 
    \in \real^{8}, \nonumber
\end{align}
where $\Frame{C_i}$ denotes the coordinate frame of the $i^{\mathrm{th}}$ camera on the sensor assembly. We used the Radial-Tangential camera model consisting of focal lengths $f_x, f_y$, centre $c_x, c_y$, radial distortion parameters $k_1, k_2$, and tangential distortion parameters $p_1, p_2$ as the camera intrinsics. Note that any other projection model could be supported in principle. The full state vector for camera calibration thus becomes,
\begin{align}
    \state = \begin{bmatrix}
        \underbrace{
            \State{FC_{1}}^{\transpose, 1}
            \dots
            \State{FC_{1}}^{\transpose, k}
        }_{\text{Reference Camera 1 Poses}}
        & \underbrace{
            \State{C_{1}C_{1}}^{\transpose}
            \dots
            \State{C_{1}C_{i}}^{\transpose}
        }_{\text{Camera Extrinsics}}
        & \underbrace{
            \State{C_1}^{\transpose}
            \dots
            \State{C_i}^{\transpose}
        }_{\text{Camera Intrinsics}}
    \end{bmatrix}^{\transpose}.
\end{align}
%The minimal error state vectors for $\State{FC_1}$ , $\State{C_1C_i}$ , and $\State{C_1C_i}$ are:
%\begin{align}
%    % Camera pose
%    \delta\boldsymbol{\chi}_{FC_{1}} &= \begin{bmatrix}
%        \delta{}_{F}\trans^{\transpose}_{FC_{1}}
%        \;\; \delta\boldsymbol{\alpha}^{\transpose}_{FC_{1}}
%    \end{bmatrix}^{\transpose} 
%    \in \real^{6}, \\
%    % Camera-camera extrinsics
%    \delta\boldsymbol{\chi}_{C_{1}C_{i}} &= \begin{bmatrix}
%        \delta{}_{C_1}\trans^{\transpose}_{C_{1}C_{i}}
%        \;\; \delta\boldsymbol{\alpha}^{\transpose}_{C_{1}C_{i}}
%    \end{bmatrix}^{\transpose} 
%    \in \real^{6}, \nonumber \\
%    % Camera intrinsics
%    \delta\boldsymbol{\chi}_{C_i} &= \begin{bmatrix}
%       \delta f_{x} 
%       \;\; \delta f_{y} 
%       \;\; \delta c_{x} 
%       \;\; \delta c_{y} 
%       \;\; \delta k_{1} 
%       \;\; \delta k_{2} 
%       \;\; \delta p_{1} 
%       \;\; \delta p_{2}
%    \end{bmatrix}^{\transpose} 
%    \in \real^{8}.
%    \nonumber
%\end{align}

\subsection{Calibration Formulation}
\label{subsec:calib_camera-formulation}

To estimate the camera calibration parameters we used a nonlinear least squares framework to minimise the cost function, $\cost_{\text{camera}}$, containing reprojection errors, $\err_{r}$, and the information matrix of the respective camera measurement, $\weight_{r}$. The cost function has the form: 
\begin{align}
    \label{eq:nbv_cost}
    \cost_{\text{camera}}(\Vec{x}) =& \enspace \frac{1}{2}\sum_{i=1}^{I} \sum_{k=1}^{K} \sum_{j \in \mathcal{J}(i, k)} 
        \err_{r}^{i,j,k^{\transpose}} \weight_{r}^{i,j,k} \err_{r}^{i,j,k}, 
\end{align}
where $i$ is the camera index, $k$ denotes the camera frame index, and $j$ denotes the fiducial target corner index. Finally, $\mathcal{J}(i, k)$ denotes the set of observable fiducial corner indices in the $i^{\text{th}}$ camera index and $k^{\text{th}}$ camera frame index.

Here, the standard reprojection error, $\err_{r}$, was used:
\begin{align}
    \err_{r}^{i, j, k} = 
        \meas^{i, j, k} - 
        \Vec{h}_{i}(
            \Tf{C_{i}}{C_{1}} \;  % Camera extrinsics
            \Tf{F}{C_{1}}^{-1} \;      % Camera pose
            \Trans{F}{F_j}{},        % Fiducial point
            \; \State{C_i}           % Intrinsics
        ),
\end{align}
whereby $\Vec{h}_{i}(\cdot)$ denotes the camera projection and distortion model. It needs as an input the fiducial corner, $\Trans{F}{F_j}{}$, camera pose, $\Tf{F}{C_{1}}$, camera extrinsics, $\Tf{C_{i}}{C_{1}}$, and camera intrinsics $\State{C_i}$. Lastly, $\meas^{i, j, k}$ is the observed fiducial corner measurement.

\subsection{Real-time Estimation}
\label{subsec:calib_camera-real_time}

Since Eq.~\eqref{eq:nbv_cost} will grow in complexity with every camera frame added, it cannot be solved in real-time as the problem size increases. We therefore adopted a fixed-lag sliding window scheme similar to~\cite{Leutenegger:etal:IJRR2014}, whereby the sliding window is bounded by marginalising out old camera poses $\state_{FC_1}$ with the Schur Complement, leading to a respective linear prior that enters the cost. Note that this is only needed for the real-time feedback to the user, and we still solve the full batch problem offline for the final calibration solution.

\subsection{Camera Calibration With Next-Best-View}
\label{subsec:calib_camera-pipeline}

\begin{figure}[!ht]
  \vspace{1em}
  \centering
  \includegraphics[width=\linewidth]{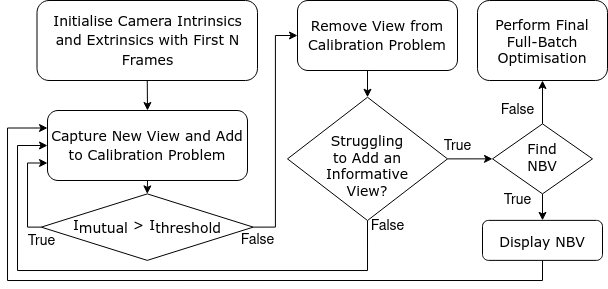}
  \caption{Camera Calibration Pipeline}
  \label{fig:nbv_pipeline}
\end{figure}

In contrast to standard full-batch camera calibration, where the calibration data is first collected and then solved as a two step process, our method takes a more integrated approach, whereby data collection and solving the calibration problem are performed incrementally, until the addition of new data is no longer informative to the camera calibration problem (see Fig.~\ref{fig:nbv_pipeline}). 

\begin{figure}[ht]
  \centering
  \includegraphics[width=0.55\linewidth]{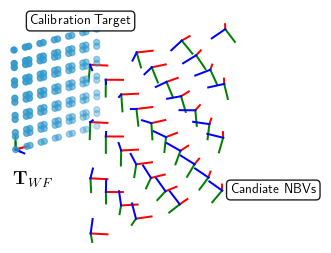}
  \caption{NBV candidate poses in-front of the calibration target}
  \label{fig:nbv_poses}
\end{figure}

First, the camera intrinsics and extrinsics are initialised with the first $N$ camera frames of a static fiducial marker of known size by minimising the cost function in Eq.~\eqref{eq:nbv_cost}. Once the camera parameters are initialised, the user is guided to maximise the calibration target measurement coverage over the image space. The information content of each camera view is evaluated using Eq.~\eqref{eq:mi}. Views which contain a MI score below the user-defined threshold, $I_{\text{threshold}}$ ($I_{\text{threshold}} = 0.2$, same as in Kalibr~\cite{Furgale:etal:IROS2013}), are removed from the calibration problem. If, however, the new candidate views are not informative enough (no new views added to the calibration problem in the last 3 frames), the calibration tool enters into ``Find Next-Best-View'' mode where it evaluates a set of possible NBVs. Similar to~\cite{Richardson:etal:IROS2013}, NBVs are pre-determined by an expert ahead of time in order to reduce the search space and make the computation feasible in real-time (see Fig.~\ref{fig:nbv_poses}). Using Eq.~\eqref{eq:mi}, the NBV is the one that has the highest mutual information. Once the NBV is determined, the calibration tool will guide the user to the NBV interactively through the graphical user-interface in capturing that view. If the mutual information of the NBV is found to be below $I_{\text{threshold}}$ the calibration tool stops capturing further measurements and proceeds to performing a final full batch optimisation to estimate the final calibration parameters.

% CAMERA-IMU EXTRINSICS CALIBRATION
\section{Camera-IMU Extrinsics Calibration}
\label{sec:nbt}

Once the camera intrinsics and extrinsics are known (from Sec.~\ref{sec:nbv}), we proceed to, without loss of generality, calibrate the extrinsics between the reference camera 1 and IMU, $\Tf{S}{C_1}$, and camera-IMU delay, $t_{d}$, of a VI-sensor.

\begin{figure}
  \vspace{1em}
  \centering
  \includegraphics[width=0.8\linewidth]{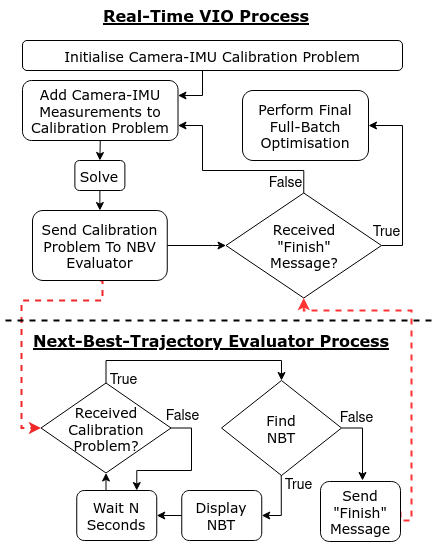}
  \caption{Camera-IMU Calibration Pipeline}
  \label{fig:nbt_pipeline}
  % \vspace{-1em}
\end{figure}

\subsection{States}
\label{subsec:calib_vi-states}

The variables to be estimated are the VI sensor pose at discrete camera frame index $k$, $\state_{WS}^{k}$, fiducial target pose in the inertial frame $\state_{WF}$, extrinsics between reference camera 1 and IMU $\state_{SC_{1}}$, and camera-IMU time delay $x_{d}$:
\begin{align}
    % IMU state
    \State{WS} &= \begin{bmatrix}
        \Trans{W}{S}{\transpose}
        \; \Quat{W}{S}^{\transpose}
        \; \Vel{W}{S}^{\transpose}
        \; \Bias{g}^{\transpose}
        \; \Bias{a}^{\transpose}
    \end{bmatrix}^{\transpose} 
    \in \real^{3} \times \mathcal{S}^{3} \times \real^{9}, 
    \nonumber \\
    % Fiducial target
    \State{WF} &= \begin{bmatrix}
        \Trans{W}{F}{\transpose}
        \;\; \Quat{W}{F}^{\transpose}
    \end{bmatrix}^{\transpose} 
    \in \real^{3} \times \mathcal{S}^{3},
    \\
    % Sensor-camera extrinsics
    \State{SC_1} &= \begin{bmatrix}
        \Trans{S}{C_1}{\transpose}
        \;\; \Quat{S}{C_1}^{\transpose}
    \end{bmatrix}^{\transpose} 
    \in \real^{3} \times \mathcal{S}^{3}, 
    \nonumber \\
    % Time delay
    x_{d} &= 
        t_{d}
    \in \real,
    \nonumber
\end{align}
where the state vector $\state_{WS}$ holds the VI sensor position in the inertial frame $\Trans{W}{S}{}$, the body orientation represented by a quaternion $\Quat{W}{S}$, the velocity expressed in the sensor frame $\vel_{WS}$, as well as the gyroscope and accelerometer biases $\Bias{g}$ and $\Bias{a}$. The state vectors $\state_{SC_1}$ and $\state_{WF}$ hold the sensor-camera relative pose and fiducial pose, respectively. The full state vector for camera-IMU calibration thus becomes,
\begin{align}
    \state = \begin{bmatrix}
        \underbrace{
            \State{WS_{1}}^{\transpose, 1}
            \dots
            \State{WS_{1}}^{\transpose, k}
        }_{\substack{\text{Sensor} \\ \text{Poses}}}
        & \underbrace{
            \State{WF}^{\transpose}
        }_{\substack{\text{Fiducial} \\ \text{Pose}}}
        & \underbrace{
            \State{SC_{1}}^{\transpose}
        }_{\substack{\text{Camera-IMU} \\ \text{Extrinsics}}}
        & \underbrace{
            x_{d}
        }_{\substack{\text{Camera-IMU} \\ \text{Time-Delay}}}
    \end{bmatrix}^{\transpose}.
\end{align}

%The minimal error state vectors for $\state_{WS}$, $\state_{SC_1}$, $\state_{WF}$, and $x_{d}$ are therefore:
%\begin{align}
%    % Error IMU state
%    \delta\boldsymbol{\chi}_{WS} &= 
%        \begin{bmatrix}
%            \delta{}_{W}\trans^{\transpose}_{WS} \enspace
%            \delta\boldsymbol{\alpha}^{\transpose}_{WS} \enspace
%            \delta\vel^{\transpose}_{WS} \enspace
%            \delta\Bias{g}^{\transpose} \enspace
%            \delta\Bias{a}^{\transpose} \enspace
%        \end{bmatrix}^{\transpose} \in \real^{15}, \nonumber \\
%    % Error sensor-camera extrinsics
%    \delta\boldsymbol{\chi}_{SC_1} &= 
%        \begin{bmatrix}
%            \delta{}_{S}\trans^{\transpose}_{SC_1} \enspace
%            \delta\boldsymbol{\alpha}^{\transpose}_{SC_1}
%        \end{bmatrix}^{\transpose} \in \real^{6}, \nonumber \\
%    % Error fiducial target
%    \delta\boldsymbol{\chi}_{WF} &= 
%        \begin{bmatrix}
%            \delta{}_{W}\trans^{\transpose}_{WF} \enspace
%            \delta\boldsymbol{\alpha}^{\transpose}_{WF}
%        \end{bmatrix}^{\transpose} \in \real^{6}, \\
%    % Error time delay
%    \delta\chi_{d} &= 
%        \delta{} t_{d}
%        \in \real.
%        \nonumber
%\end{align}

\subsection{Calibration Formulation}
\label{subsec:calib_vi-formulation}

Similar to Sec.~\ref{sec:nbv}, we seek to formulate the VI calibration problem as one joint nonlinear-optimisation of a cost function $\cost_{\text{imu-cam}}(\state)$ containing both (weighted) reprojection errors $\err_{r}$ and (weighted) temporal error term from the IMU $\err_{s}$:
\begin{align}
    \label{eq:nbt_cost}
    \cost_{\text{imu-cam}}(\state) 
    =& \underbrace{
        \dfrac{1}{2} 
        \sum_{i=1}^{I} \sum_{k=1}^{K} \sum_{j \in \mathcal{J}(i, k)} 
        \err_{r}^{i,j,k^{\transpose}} \weight_{r}^{i,j,k} \err_{r}^{i,j,k}
    }_{\text{visual}}
    \\ \nonumber
    &+ \underbrace{
        \dfrac{1}{2}
        \sum_{k=1}^{K-1}
        \err_{s}^{k^{\transpose}} \weight_{s}^{k} \err_{s}^{k}
    }_{\text{inertial}},
\end{align}
where $i$ is the camera index of the VI sensor, $k$ denotes the camera frame index, and $j$ denotes the fiducial target corner index. The set $\mathcal{J}(i, k)$ represents the indices of fiducial target corners observed in the $k^{\text{th}}$ frame and the $i^{\text{th}}$ camera.

The reprojection error was used to estimate the camera-IMU extrinsics $\Tf{S}{C_1}$, sensor pose in the world frame $\Tf{W}{S}$ and fiducial target in the world frame $\Tf{W}{F}$:
\begin{equation}
    \err_{r} = \meas^{i,j,k} - 
        \mmodel_{i}(
            \Tf{C_1}{C_i}^{-1}     % Camera-camera extrinsics
            \Tf{S}{C_1}^{-1}       % IMU-camera extrinsics 
            \Tf{S}{W}^{k}          % Sensor pose in world frame
            \Tf{W}{F}              % Fiducial pose in world frame
            \Trans{F}{F_{j}}{},    % Fiducial corner
            \;
            \state_{C_i}
        ),
\end{equation}
where $\mmodel_{i}(\cdot)$ denotes the $i^{\text{th}}$ camera projection model which includes distortion, $\Trans{F}{F_j}{}$ denotes the $j^{\text{th}}$ fiducial target corner point and $\meas^{i,j,k}$ denotes the corresponding measurement seen in camera $i$ and image frame $k$ in image coordinates. The camera-intrinsics $\state_{C_i}$ and camera-extrinsics $\Tf{C_1}{C_i}$ estimated in Sec.~\ref{sec:nbv} are fixed.

The fiducial target in the world frame $\Tf{W}{F}$ is first initialised using initial measurements from the IMU and camera assuming low acceleration, where the measured acceleration vector corresponds to (inverse) acceleration due to gravity--yielding the camera pose $\Tf{W}{C_i}$. Without loss of generality, we set the camera position and yaw around the world-z axis to zero. Next, the relative pose between the fiducial target and the $i^{\text{th}}$ camera, $\Tf{F}{C_i}$, is computed with fiducial corner measurements using 3D-2D RANSAC and bundle adjustment, after which we can compose $\Tf{W}{F} = \Tf{W}{C_i} \Tf{C_i}{F}$.

For the IMU error term, we adopted the pre-integration scheme in~\cite{Forster:etal:RSS2015}, where the error is the difference between the predicted relative state and the actual relative state, with the exception of orientation, where a simple multiplicative minimal error was used:
\begin{equation}
    \error_{S}^{k} (\state_{S}^{k}, \state_{S}^{k + 1}, \meas_{S}^{k}) =
        \begin{bmatrix}
            % Position
            \EstBodyTrans{W}{S}^{k, k + 1}(t_{d}) - \BodyTrans{W}{S}^{k, k + 1} \\ %\vspace{0.5em} \\
            % Orientation
            2 \begin{bmatrix}
                \Quat{}{S}^{k, k + 1} \otimes
                \EstQuat{}{S}^{k, k + 1}(t_{d})
            \end{bmatrix}_{1:3} \\ % \vspace{0.5em}\\
            % Speed and biases
            {}_{S}\hat{\vel}^{k, k + 1}(t_{d}) - {}_{S}\vel^{k, k + 1} \\ % \vspace{0.5em} \\
            \hat{\bias}_{g}^{k + 1}(t_{d}) - \bias_{g}^{k + 1}  \\ % \vspace{0.5em} \\
            \hat{\bias}_{a}^{k + 1}(t_{d}) - \bias_{a}^{k + 1}
        \end{bmatrix} \in \real^{15}.
\end{equation}
In addition to estimating the relative state, we further include the camera-IMU time delay scalar $t_{d}$. Since it is only a 1 dimensional parameter, the $15 \times 1$ Jacobian was obtained through the central finite difference by perturbing the IMU timestamps.

\subsection{Real-time Estimation}
\label{subsec:calib_vi-realtime}

To keep the problem in Eq.~\eqref{eq:nbt_cost} bounded for real-time operation, we used the same approach as in Sec.~\ref{subsec:calib_camera-real_time} and adopted a fixed-lag sliding window scheme, marginalising out old
sensor poses $\Tf{W}{S}$, velocities $\Vel{W}{S}^{}$, accelerometer biases $\bias_{\accel}$ and gyroscope biases $\bias_{g}$. A full batch optimisation using all measurements will be performed to obtain the final calibration solution. The camera-IMU time delay parameter is fixed during online guidance, and estimated in the final full batch optimisation.

\subsection{Next-Best-Trajectories}
\label{subsec:calib_vi-nbts}

\begin{figure}
\begin{center}
    \begin{subfigure}[b]{0.35\linewidth}
        \includegraphics[width=\linewidth]{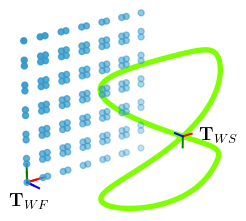}
        \caption{Vertical Figure-8}
    \end{subfigure}
    \begin{subfigure}[b]{0.35\linewidth}
        \includegraphics[width=\linewidth]{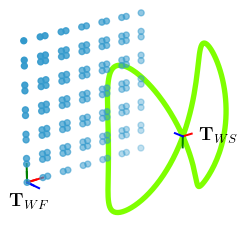}
        \caption{Horizontal Figure-8}
    \end{subfigure}

    \begin{subfigure}[b]{0.35\linewidth}
        \includegraphics[width=\linewidth]{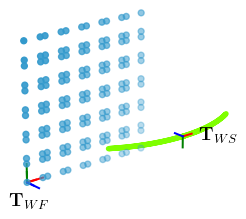}
        \caption{Horizontal Pan}
    \end{subfigure}
    \begin{subfigure}[b]{0.35\linewidth}
        \includegraphics[width=\linewidth]{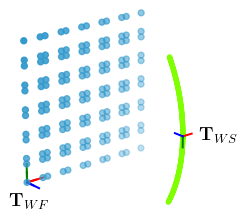}
        \caption{Vertical Pan}
    \end{subfigure}
    
    \begin{subfigure}[b]{0.35\linewidth}
        \includegraphics[width=\linewidth]{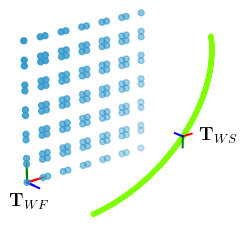}
        \caption{Diagonal 0}
    \end{subfigure}
    \begin{subfigure}[b]{0.35\linewidth}
        \includegraphics[width=\linewidth]{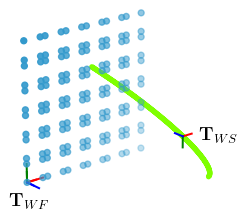}
        \caption{Diagonal 1}
    \end{subfigure}
    \caption{Next-Best-Trajectory (NBT) candidates in-front of a calibration target}
    \label{fig:nbts}
\end{center}
\end{figure}

Similar to~\cite{Nobre:Heckman:IJRR2019}, given our goal is to provide intuitive, easy and real-time feedback for a non-expert user to calibrate the VI-sensor, we discretized the continuous search space and used the results of~\cite{Yang:etal:RAL2019} to design 6 non-degenerate NBTs that are computationally feasible in real-time, easy to display and followed by the user (see Fig.~\ref{fig:nbts}). Our NBTs are observable as the fisher-information matrix has to be invertible in order to evaluate the information gain~\cite{Nobre:Heckman:IJRR2019}.

Inspired by the Lissajous curve equations, each NBT is parameterised as:
\begin{align}
    \label{eq:lissajous-positions}
    x &= w_{\text{traj}} \sin(at + \delta) + 0.5 w_{\text{calib}}, \nonumber \\
    y &= h_{\text{traj}} \cos(bt) + 0.5 h_{\text{calib}}, \\
    z &= \sqrt{d_{\text{nbt}} - x^{2} - y^{2}}, \nonumber
\end{align}
where $x$, $y$ and $z$ are the trajectory positions relative to the fiducial target frame $\Frame{F}$ to form $\Trans{F}{S}{}$, $d_{\text{nbt}}$ is the distance away from the fiducial target center, $w_{\text{traj}}$ and $h_{\text{traj}}$ are the trajectory max width and height, $w_{\text{calib}}$ and $h_{\text{calib}}$ are the fiducial target width and height, $\delta$ represents the phase angle offset, and finally $a$ and $b$ are constants that determine the shape of the trajectory (e.g. a ratio of $\frac{a}{b} = 2$ forms a figure of 8). Finally, the sensor's orientation are parameterised as Euler angles and designed such that it is always pointing towards the center of the calibration target:
\begin{align}
    \label{eq:lissajous-orientations}
   \phi &= \phi_{\text{bound}} \sin(2 \pi t) + \pi, \nonumber \\
   \theta &= \theta_{\text{bound}} \sin(2 \pi t), \\
   \psi &= 0.0, \nonumber
\end{align}
where $\phi$, $\theta$ and $\psi$ are Euler angles around the x, y and z-axis to form $\Rot{F}{S}$, respectively, and $\phi_{\text{bound}}$ and $\theta_{\text{bound}}$ are the maximum rotation around x and y-axis, respectively.

To ensure the velocity and angular velocity are realistic, we parameterise $t$ in Eq.~\eqref{eq:lissajous-positions} and Eq.~\eqref{eq:lissajous-orientations} as a function of $k$ between $[0, t_{\text{nbt}}]$ such that the first derivative of both equations, velocity and angular velocity, start and end at 0,
$t(k) = \sin^{2}\left(\pi k/ 2 \; t_{\text{nbt}}\right)$,
% \begin{equation}
%     \label{eq:lissajous-time}
%     t(k) = \sin^{2}\left(\dfrac{\pi k}{2\; t_{\text{nbt}}}\right),
% \end{equation}
where $t_{\text{nbt}}$ is the time to complete a NBT. Differentiating both Eq.~\eqref{eq:lissajous-positions} and Eq.~\eqref{eq:lissajous-orientations}
enables us to simulate the camera and IMU measurements for evaluating NBTs using Eq.~\eqref{eq:mi}.

\subsection{Camera-IMU Calibration With Next-Best-Trajectory}
\label{subsec:calib_vi-pipeline}

The Camera-IMU calibration begins with two separate processes running in parallel, a real-time VI estimator solving Eq.~\eqref{eq:nbt_cost} and a NBT evaluator (see Fig.~\ref{fig:nbt_pipeline}). The real-time VI camera parameters are initialised using the parameters optimised in Section.~\ref{sec:nbv} and are fixed throughout. The fiducial pose, $\Tf{W}{F}$ and camera-IMU extrinsics, $\Tf{S}{C_1}$ on the other hand are initialised by solving Eq.~\eqref{eq:nbt_cost} with the first $N$ camera frames, and IMU measurements between the first and last camera frame timestamps. 

As the real-time VI  estimator is solving the camera-IMU calibration problem, it periodically sends the calibration problem data to the NBT evaluator process. The NBT evaluator in turn would use the data to evaluate the MI of a set of pre-defined NBTs (Section.~\ref{subsec:calib_vi-nbts}) using Eq.~\eqref{eq:mi}, find the NBT with the highest MI and guide the user in executing the NBT in order to render the calibration parameters optimally observable, i.e.\ reducing the expected uncertainty on the estimated camera-IMU extrinsics. If none of the candidate NBTs satisfies $I_{\text{mutual}} > I_{\text{threshold}}$ ($I_{\text{threshold}} = 0.2$) then the NBT evaluator sends a ``finish'' message to communicate to the real-time VI process that it should proceed to perform a final full-batch calibration.

% Experiments
\section{Experiments}
\label{sec:experiments}

To evaluate our method, we conducted two sets of experiments. First, we evaluated our calibration pipeline in offline mode with the EuRoC~\cite{Burri:etal:IJRR2016} dataset to verify our calibration \emph{accuracy} is competitive against that of Kalibr's \textit{without the interactive component} of our system, and despite different approaches to solving the camera-IMU calibration problem, where Kalibr uses a continuous time full-batch optimisation in contrast to our method which uses a discrete time full-batch optimisation. This is \textit{independent} of our contributions regarding interactivity. With this we wanted to highlight our calibration tool without interactivity is at least as good as Kalibr.

Since the main motivation in this work is to provide non-experts with good calibration results for VIO/VI-SLAM systems, we further conducted experiments involving a small batch of graduate students to prove that our system can \emph{efficiently} and \emph{reliably} calibrate VI sensors, achieving superior performance for existing VIO and VI-SLAM systems. To compare the calibrations we used them in ORBSLAM3~\cite{Campos:etal:TRO2021} and evaluated the accuracy using the evaluation scheme of~\cite{Sturm:etal:IROS2012} with RMSE Aboslute Trajectory Error (ATE) by aligning the estimated trajectory with the ground-truth.

All experiments were conducted on a Lenovo P52 Thinkpad laptop containing an Intel Core i7-8750H CPU at 2.2 Ghz with 16GB of memory running Ubuntu 20.04 and ROS Melodic. The experiments with the graduate students were conducted with the aim of calibrating an Intel RealSense D435i which contains a stereo IR global shutter depth sensor, a monocular RGB rolling shutter sensor, and additionally an IMU sensor running at 15Hz, 15Hz and 400Hz respectively. For our purposes, we \textit{do not} use the RGB rolling shutter sensor. We have instead disabled the IR projector and used the stereo IR global shutter depth sensors as a standard gray-scale stereo camera. 

During the camera and VI calibrations the default settings for Kalibr~\cite{Furgale:etal:IROS2013} were used to generate their results, whereas in our method we used Cauchy loss ($s = 1.5$) on the reprojection errors and a fixed-lag smoothing window size of 10 and 3 for the camera calibration and camera-IMU calibration stages respectively. The IMU parameters used for the camera-IMU calibration are: $\sigma_{a} = 2.52 \times10^{-2} \frac{\mathrm{m}}{\mathrm{s}^2}\frac{1}{\sqrt{\mathrm{Hz}}}$ for the accelerometer noise density, $\sigma_{ba} = 4.41 \times 10^{-3} \frac{\mathrm{m}}{s^3}\frac{1}{\sqrt{\mathrm{Hz}}}$ for the accelerometer drift noise density, $\sigma_{g} = 2.78 \times10^{-3} \frac{\mathrm{rad}}{\mathrm{s}}\frac{1}{\sqrt{\mathrm{Hz}}}$ for gyroscope noise density, and $\sigma_{bg} = 1.65 \times10^{-5} \frac{\mathrm{rad}}{s^2}\frac{1}{\sqrt{\mathrm{Hz}}}$ gyroscope drift noise density.

% \begin{table}[!ht]
% \centering
% \caption{IMU parameters}
% \begin{tabular}{l l l}
% \hline
% & \\[-1.5ex]
% \textbf{Parameter} & 
% \textbf{Value} & 
% \textbf{Unit} \\
% & \\[-1.5ex]
% \hline
% \\[-1.5ex]
% Accelerometer Noise Density $\sigma_{a}$          
%     & $2.52 \times10^{-2} $ 
%     & $\dfrac{m}{s^2}\dfrac{1}{\sqrt{Hz}}$ \\ [2ex]
% Accelerometer Drift Noise Density $\sigma_{ba}$   
%     & $4.41 \times10^{-3} $ 
%     & $\dfrac{m}{s^3}\dfrac{1}{\sqrt{Hz}}$ \\ [2ex]
% Gyroscope Noise Density $\sigma_{g}$              
%     & $2.78 \times10^{-3} $ 
%     & $\dfrac{rad}{s}\dfrac{1}{\sqrt{Hz}}$ \\ [2ex]
% Gyroscope Drift Noise Density $\sigma_{bg}$       
%     & $1.65 \times10^{-5} $ 
%     & $\dfrac{rad}{s^2}\dfrac{1}{\sqrt{Hz}}$ \\
% & \\[-1.5ex]
% \hline
% \end{tabular}
% \label{tbl:imu_params}
% \end{table}

\subsection{Calibration Results on EuRoC Dataset}

To assess our approach in offline mode, we used the calibration sequences from the EuRoC dataset~\cite{Burri:etal:IJRR2016} to calibrate the VI-sensor. The calibration process is split into two stages. First, the camera intrinsics and camera extrinsics are estimated. Then in the second stage, only the camera-IMU extrinsics and time-delay are estimated, with the camera intrinsics and camera extrinsics estimated in the first phase fixed.

\begin{figure}[htb]
  \centering
  \includegraphics[width=\linewidth]{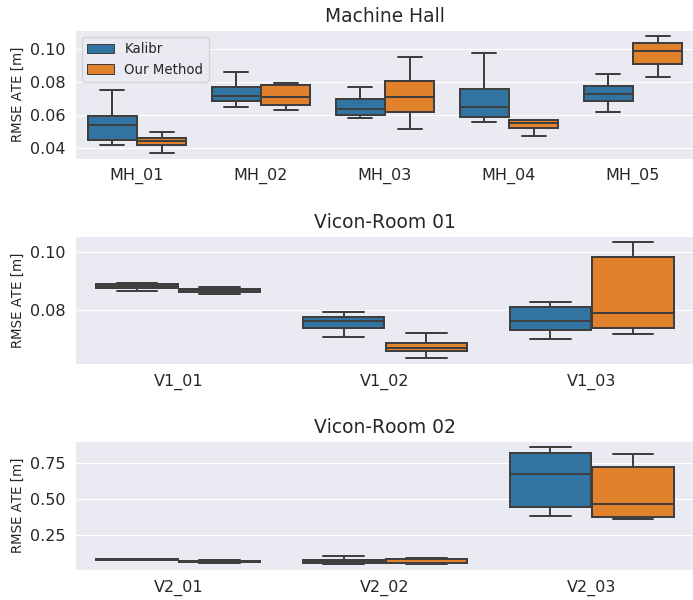}
  \caption{Comparison of ORBSLAM3 using calibrations from Kalibr and ours in Stereo-VO mode on EuRoC Dataset}
  \label{fig:euroc-vo}
  % \vspace{-1em}
\end{figure}

\begin{figure}[htb]
  \centering
  \includegraphics[width=\linewidth]{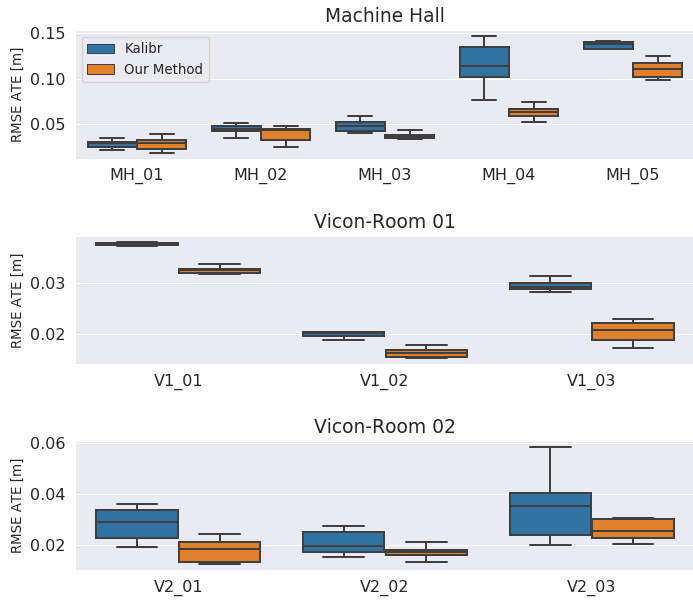}
  \caption{Comparison of ORBSLAM3 using calibrations from Kalibr and ours in Stereo-VIO mode on EuRoC Dataset}
  \label{fig:euroc-vio}
\end{figure}

The results show comparable calibration reprojection errors, where in the camera calibration stage our method obtained an RMSE reprojection error of $0.6042$ pixels compared to Kalibr's $0.6087$ pixels, and in the camera-IMU calibration stage the RMSE reprojection errors are $0.5569$ pixels and $0.5775$ pixels for our method and Kalibr respectively. Fig.~\ref{fig:euroc-vo} and Fig.~\ref{fig:euroc-vio} report RMSE ATE after running ORBSLAM3~\cite{Campos:etal:TRO2021} on the EuRoC dataset sequences 10 times in Stereo-VO mode and Stereo-VIO mode, respectively. We did not change the ORB-SLAM3 EuRoC configuration that was orginally tuned for Kalibr calibration parameters. Both figures show that the calibrations produced by our method yielded better results on most sequences in Stereo-VO mode and all sequences in VIO mode, compared to Kalibr.

To verify our camera-IMU time delay estimation, we assumed the EuRoC dataset~\cite{Burri:etal:IJRR2016} has a camera-IMU time delay of $\approx 7\mu s$, as reported in~\cite{Nikolic:etal:ICRA2014}, and perturbed the \texttt{imu\_april} IMU timestamps with 100ms, 10ms and 1ms time offsets. With our offline camera-IMU calibration \textit{without interactivity} we were able to recover the time offsets $100ms$, $9.97ms$ and $0.987ms$ respectively, thus showing that our offline camera-IMU calibrator is capable of accurately estimating the camera-IMU time delay.

\subsection{Trials with Graduate Students}

To evaluate our calibration method, we conducted a series of tests involving 16 graduate students to measure the effectiveness of our approach compared to the state-of-the-art calibrator, Kalibr~\cite{Furgale:etal:IROS2013}. Our test-subjects were postgraduate students at Imperial College London. Of the 16 students, 4 reported some previous experience with camera calibration, and only 2 reported some previous experience with camera-IMU calibration. Each participant was asked to calibrate the same Intel RealSense D435i sensor by first collecting two calibration sequences for Kalibr (one for camera calibration and the second for camera-IMU calibration), and then another two with our calibration method.

Because we do not have ground truth for the calibration parameters, we evaluated the estimated calibration parameters by applying them in ORBSLAM3~\cite{Campos:etal:TRO2021} running in odometry mode (with loop-closure switched off) on 10 custom-collected Vicon room sequences where ground-truth poses were recorded with various motions. 

\begin{figure}[tb]
  \centering
  \includegraphics[width=\linewidth]{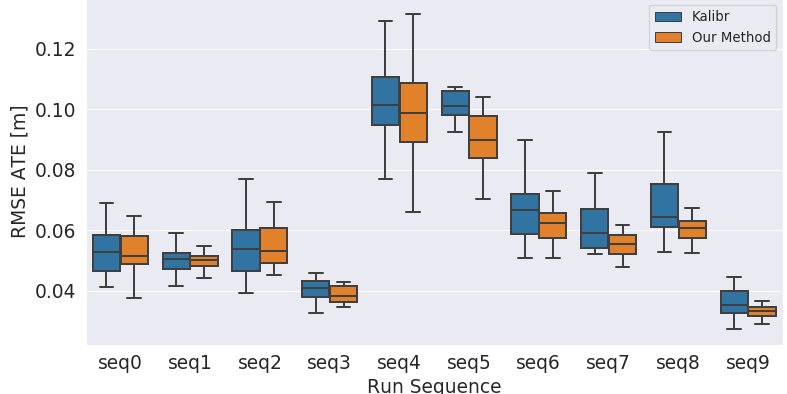}
  \caption{Comparing calibrations by graduate students across 10 different evaluation VICON room sequences by running ORBSLAM3 in odometry mode}
  \label{fig:odom-boxplot}
\end{figure}

\begin{figure}[htb]
  \centering
  \includegraphics[width=\linewidth]{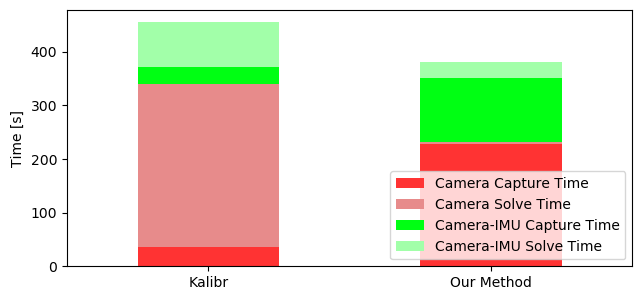}
  \caption{Median total calibration time per person in seconds}
  \label{fig:timings-stacked}
\end{figure}

Our study shows that novices who have little to no experience in calibrating a VI sensor can obtain better calibrations using our approach compared to Kalibr. Out of 10 Vicon room sequences, the RMSE ATE error is lowest across all sequences using calibration parameters obtained through our method (see Fig.~\ref{fig:odom-boxplot}). Our calibration parameters also yielded overall smaller RMSE ATE variances, showing more consistent and reliable odometry accuracy, regardless of the experience of calibration users. The estimated camera-IMU time delay with our method is $3.07 \pm 0.932 ms$, and Kalibr's estimate is $4.81 \pm 0.981 ms$. Since ground-truth is not available we can only conclude our method is more consistent compared to Kalibr's result. The break-down of the median total time taken to calibrate the VI sensor between Kalibr and our method is shown in Fig.~\ref{fig:timings-stacked}, where our method's median is 381.11 seconds compared to Kalibr's 455.44 seconds. %It is worth noting even though our method spends more time collecting camera data, the number of \textit{informative} views used by both methods are similar where Kalibr used on average 117 views compared to 114 views with our method.

% The median total time taken to calibrate the VI sensor with our method is 381.11 seconds compared to Kalibr's 455.44 seconds. It is worth noting that even though our method spent more time collecting camera calibration data, the actual number of \textit{informative} views used by both methods are similar where Kalibr uses on average 117 views compared to 114 views with our method.

\begin{figure}[!htb]
    \begin{subfigure}[b]{0.45\linewidth}
        \includegraphics[width=\linewidth]{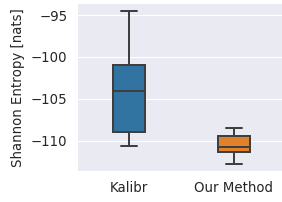}
        \caption{Camera Calibration}
        \label{fig:calib_camera-entropy}
    \end{subfigure}
    \hfill
    \begin{subfigure}[b]{0.45\linewidth}
        \includegraphics[width=\linewidth]{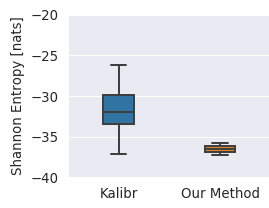}
        \caption{Camera-IMU Calibration}
        \label{fig:calib_imu-entropy}
    \end{subfigure}
    \caption{Shannon Entropy of camera and camera-IMU calibration across all calibrations by graduate students}
    % \vspace{-0.8em}
\end{figure}

In addition to showing that our method yields better SLAM results and calibrations faster, by inspecting the Shannon Entropy of the calibration parameters, a metric used to measure the uncertainty of information content~\cite{Cover:Thomas:book2006}, we also observe a lower entropy (more certainty) with our method compared to Kalibr (see Fig.~\ref{fig:calib_camera-entropy} and Fig.~\ref{fig:calib_imu-entropy}). This means that our method can successfully guide a novice to collect a more informative calibration dataset for a good calibration.

\section{CONCLUSIONS}

The success of SoTA computer-vision and state-estimation algorithms often hinges on good VI calibrations. However, collecting high-quality VI calibration data is not trivial, especially since most existing calibration tools do not provide an interactive live feedback to the user which ultimately increases the risk of poor calibrations. In this work, we have introduced a novel visual-inertial calibration guidance system to provide real-time NBV and NBT suggestions to guide users in collecting informative calibration data. It achieves competitive calibration results against the SoTA offline calibrator, Kalibr~\cite{Furgale:etal:IROS2013}, and produces faster, more accurate and more reliable calibrations for existing SoTA visual and VI SLAM systems, even when used by novices. 

%%%%%%%%%%%%%%%%%%%%%%%%%%%%%%%%%%%%%%%%%%%%%%%%%%%%%%%%%%%%%%%%%%%%%%%%%%%%%%%%
\addtolength{\textheight}{-12cm}  % This command serves to balance the column lengths

% ACKNOWLEDGEMENT
\section*{ACKNOWLEDGMENT}

We thank members from the Smart Robotics Lab, Robot Learning Lab, Dyson Robotics Lab and Adaptive and Intelligent Robots Lab for participating in experiments, Ying Xu for graphic design, and especially Sotiris Papatheodorou for his fruitful advice in this project. This research is supported by Imperial College London, Technical University of Munich, EPSRC grant ORCA Stream B - Towards Resident Robots, and the EPSRC grant Aerial ABM EP/N018494/1.

\bibliographystyle{unsrt}
\bibliography{abbr_short,robotvision}
\end{document}